\documentclass[conference]{IEEEtran}
\IEEEoverridecommandlockouts
\usepackage{cite}
\usepackage{amsmath,amssymb,amsfonts}
\usepackage{graphicx}
\usepackage{textcomp}
\usepackage{xcolor}
\usepackage{cleveref}
\usepackage{url}
\crefname{section}{§}{§§}
\Crefname{section}{§}{§§}
\crefname{table}{Table}{Tables}
\Crefname{table}{Table}{Tables}
\crefname{figure}{Figure}{Figures}
\Crefname{figure}{Figure}{Figures}
\crefname{equation}{Equation}{Equations}
\Crefname{equation}{Equation}{Equations}
\crefname{algorithm}{Algorithm}{Algorithms}
\Crefname{algorithm}{Algorithm}{Algorithms}

\usepackage{algorithm} %
\usepackage{algorithmic} %

\definecolor{shapecolor}{rgb}{0.6,0.1,0.0}
\newcommand{\name}{NetMamba}%

\definecolor{commentcolor}{rgb}{0.5,0.5,0.5}

\usepackage{enumitem}
\usepackage{booktabs}
\usepackage{multirow}
\usepackage{threeparttable}
\usepackage{colortbl}
\usepackage{pifont}
\usepackage{array}

\def\BibTeX{{\rm B\kern-.05em{\sc i\kern-.025em b}\kern-.08em
    T\kern-.1667em\lower.7ex\hbox{E}\kern-.125emX}}

\IEEEoverridecommandlockouts\IEEEpubid{\makebox[\columnwidth]{ 979-8-3503-5171-2/24/\$31.00 $\copyright$2024 IEEE \hfill}\hspace{\columnsep}\makebox[\columnwidth]{ }}
\begin{document}

\title{\name{}: Efficient Network Traffic Classification via Pre-training Unidirectional Mamba}

\author{
\IEEEauthorblockN{Tongze Wang\textsuperscript{1}, Xiaohui Xie\textsuperscript{2*} 
\thanks{This work is supported by the NSFC Project under Grant 62132009, Grant 62221003 and Grant 62394322.}
\thanks{\textsuperscript{*} Corresponding Authors: Xiaohui Xie and Yong Cui}, Wenduo Wang\textsuperscript{2}, Chuyi Wang\textsuperscript{2}, Youjian Zhao\textsuperscript{2}, Yong Cui\textsuperscript{2*}}
\IEEEauthorblockA{\textsuperscript{1}\textit{Institute for Network Sciences and Cyberspace, Tsinghua University}\\
\textsuperscript{2}\textit{Department of Computer Science and Technology, Tsinghua University}}
}

\maketitle

\begin{abstract}
Network traffic classification is a crucial research area aiming to enhance service quality, streamline network management, and bolster cybersecurity. To address the growing complexity of transmission encryption techniques, various machine learning and deep learning methods have been proposed. However, existing approaches face two main challenges. Firstly, they struggle with model inefficiency due to the quadratic complexity of the widely used Transformer architecture. Secondly, they suffer from inadequate traffic representation because of discarding important byte information while retaining unwanted biases.
To address these challenges, we propose \name{}, an efficient linear-time state space model equipped with a comprehensive traffic representation scheme. 
We adopt a specially selected and improved unidirectional Mamba architecture for the networking field, instead of the Transformer, to address efficiency issues. 
In addition, we design a traffic representation scheme to extract valid information from massive traffic data while removing biased information.
Evaluation experiments on six public datasets encompassing three main classification tasks showcase \name{}'s superior classification performance compared to state-of-the-art baselines. 
It achieves accuracy rates exceeding 90\%, with some surpassing 99\%, across all tasks.
Additionally, \name{} demonstrates excellent efficiency, improving inference speed by up to 60 times while maintaining comparably low memory usage. Furthermore, \name{} exhibits superior few-shot learning abilities, achieving better classification performance with fewer labeled data. 
To the best of our knowledge, \name{} is the first model to tailor the Mamba architecture for networking.

\end{abstract}


\section{Introduction}
Network traffic classification, which aims to identify potential threats within traffic or classify the category of traffic originating from different applications or services, has become an increasingly vital research area.
This is crucial for ensuring cybersecurity, improving service quality and user experience, and enabling efficient network management. 
However, the widespread adoption of encryption techniques (e.g., TLS) and anonymous network technologies (e.g., VPN, Tor) has made the accurate analysis of complex traffic more challenging.

Researchers have proposed numerous approaches to address this issue, showing promising results yet facing severe limitations.
%transitioning from conventional machine learning methodologies to more sophisticated deep learning techniques. 
Conventional machine learning methods~\cite{hayes2016k, van2020flowprint, taylor2017robust}, primarily relying on manually engineered features or statistical attributes, often fail to capture accurate traffic representations due to the absence of raw traffic data. 
In contrast, deep learning approaches~\cite{liu2019fs, lotfollahi2020deep, zhang2023tfe} automatically extract features from raw byte-level data, leading to enhanced traffic classification capabilities. 
Nonetheless, these deep learning methods necessitate extensive labeled datasets, rendering the models susceptible to biases and impeding their adaptability to novel data distributions.

Recently, pre-training has emerged as a prevalent model training paradigm in natural language processing~(NLP)~\cite{devlin2018bert} and computer vision~(CV)~\cite{he2022masked}. 
Motivated by this trend, several Transformer-based pre-trained traffic models~\cite{lin2022bert, zhao2023yet, wang2024lens} have been developed to learn generic traffic representations from extensive unlabeled data and then fine-tune for specific downstream tasks using limited labeled traffic data. However, these existing models face two significant challenges: 1) Limited Model Efficiency: state-of-the-art methods in traffic analysis primarily use Transformer architecture, which employs a quadratic self-attention mechanism to calculate correlations within a sequence. This leads to substantial computational and memory costs on long sequences\cite{zhu2024vision, qu2024trafficgpt}. Consequently, these models are unsuitable for real-time online traffic classification and cannot operate efficiently with the limited resources of typical network devices. 2) Inadequate Traffic Representation: current methodologies fail to adequately and accurately represent raw traffic data due to discarding crucial byte information and preserving unwanted biases. As a result, these unreliable schemes impair classification performance or even cause model failure in complex traffic scenarios.

To address these challenges, we propose \name{}, an efficient linear-time state space model equipped with a comprehensive traffic representation scheme, aiming to accurately perform network traffic classification tasks with higher inference speed and lower memory usage.

To improve model efficiency, we use the Mamba architecture for the model backbone instead of the Transformer.
Mamba\cite{gu2023mamba}, a liner-time state space model for sequence modeling, has achieved notable success across various domains, including natural language processing~\cite{he2024densemamba}, computer vision~\cite{zhu2024vision} and graph understanding~\cite{wang2024graph}. 
This suggests promising potential for applying Mamba to the network domain. However, adapting Mamba for efficient and robust network traffic analysis requires selecting the appropriate architecture from the existing heterogeneous and complex Mamba variants. 
By carefully testing different variants of Mamba, we found that the original unidirectional Mamba\cite{gu2023mamba}, without omnidirectional scans or redundant blocks, is well-suited for efficiently learning latent patterns within sequential network traffic. 
To further enhance the model's performance and robustness, we incorporate positional embeddings and pre-training strategies specially designed for networking.

To enhance traffic representation, we design a more comprehensive and reliable scheme. This scheme retains valuable packet content within both headers and payloads while eliminating unwanted biases through various methods, including packet anonymizing, byte allocation balancing and stride-based data cutting, thereby improving traffic classification capabilities.

% Based on the aforementioned designs, we propose \name{}, a novel pre-trained unidirectional state space model for efficient and robust network traffic classification. 
% Benefiting from the linear-time complexity and memory optimizations of Mamba~\cite{gu2023mamba}, the efficiency of \name{} is secured. 
% Additionally, we design a more comprehensive traffic representation scheme that incorporates richer packet information while reducing bias, thereby enhancing traffic classification capabilities. 
Specifically, \name{} initially extracts hierarchical flow information from raw traffic and converts it into a stride sequence, which serves as the model's input. 
Subsequently, \name{} undergoes self-supervised pre-training on large unlabeled datasets using a masked autoencoder structure, which is designed to learn generic representations of traffic data through reconstructing masked strides. 
Finally, the decoder is replaced with a multi-layer perceptron head, and \name{} is fine-tuned on limited labeled data to refine traffic representations and adapt to downstream traffic classification tasks.
% \textcolor{red}{}
Extensive experiments conducted on publicly available datasets demonstrate the effectiveness and efficiency of \name{}. In all classification tasks, \name{} consistently achieves accuracy rates above 90\%, with some exceeding 99\%. Compared to existing baselines, it improves inference speed by up to 60 times while maintaining low GPU memory usage. Furthermore, \name{} exhibits superior few-shot learning capabilities in comparison to other pre-training models, achieving better performance with fewer labeled data.

In summary, our work makes the following contributions:
\begin{enumerate}[label=(\arabic*)]
    \item We propose \name{}, the first state space model specifically designed for network traffic classification. Compared to existing Transformer-based methods, \name{} demonstrates superior performance and inference efficiency.
    \item We develop a comprehensive representation scheme for network traffic data that preserves valuable traffic characteristics while eliminating unwanted biases.
    \item We conduct extensive experiments across a range of traffic classification tasks. An overall comparison, along with detailed evaluations—encompassing ablation studies, efficiency analyses, and few-shot learning investigations—is provided. These insights could illuminate paths for future research. 
    Additionally, the code of \name{} is publicly available~\footnote{\url{https://github.com/wangtz19/NetMamba}}. 
\end{enumerate}
\section{Related Work}
\subsection{Transformer-based Traffic Classification}
% \textcolor{red}{traffic analysis, traffic understanding or traffic classification?}
Due to its highly parallel architecture and robust sequence modeling abilities, Transformer has gained significant popularity and is extensively used for traffic understanding and generation tasks. For instance, MTT~\cite{zheng2022mtt} employs a multi‑task Transformer trained on truncated packet byte sequences to analyze traffic features in a supervised way. Recognizing the challenges associated with data annotation, MT-FlowFormer~\cite{zhao2022mt} introduces a Transformer-based semi-supervised framework for data augmentation and model improvement. 

To leverage unlabeled data effectively, several pre-trained models have been proposed. Inspired by BERT's pre-training methodology in natural language processing, PERT~\cite{he2020pert} and ET-BERT~\cite{lin2022bert} process raw traffic bytes using tokenization, apply masked language modeling to learn traffic representations, and fine-tune the models for downstream tasks. Similarly, YaTC~\cite{zhao2023yet} and FlowMAE~\cite{hang2023flow} adopt the widely-used MAE pre-training approach from computer vision, which involves patch splitting for byte matrices, capturing traffic correlations through masked patch reconstruction, and subsequent fine-tuning. 

Given the global interest in large language models, pre-trained traffic foundation models such as NetGPT~\cite{meng2023netgpt} and Lens~\cite{wang2024lens} have been developed to address traffic analysis and generation simultaneously.
% \textcolor{red}{However, transformer-based models...}
However, Transformer-based models face computational and memory inefficiencies because of the quadratic complexity of their core self-attention mechanism. This necessitates a more efficient and effective solution for online traffic classification.

\subsection{Mamba-based Representation Learning}
Representation learning is a branch of machine learning concerned with automatically learning and extracting meaningful representations or features from raw data. Since the advent of Mamba, an efficient and effective sequence model, numerous Mamba variants have emerged to enhance representation learning across diverse domain-specific data formats. For instance, in the realm of vision tasks requiring spatial awareness, custom-designed scan architectures like Vim~\cite{zhu2024vision} and VMamba~\cite{liu2024vmamba} have been developed. In the domain of language modeling, DenseMamba~\cite{he2024densemamba} improves upon the original SSM by incorporating dense internal connections to boost performance. Handling graph data necessitates specialized solutions such as Graph-Mamba~\cite{wang2024graph} and STG-Mamba~\cite{li2024stg}, each employing tailored graph-specific selection mechanisms. Furthermore, various Mamba variants have proven effective in domains like signal processing~\cite{li2024spmamba}, point cloud analysis~\cite{liang2024pointmamba}, and multi-modal learning~\cite{qiao2024vl}.

However, to date, there are no reports of Mamba's successful application in network traffic classification, highlighting the need for our research in this area.

\subsection{Traffic Representation Schemes}

In real-world scenarios, massive raw network traffic encompasses a wide range of data categories that vary in upper applications, carried protocols, or transmission purposes. Therefore, a robust representation scheme with appropriate granularity is crucial for accurate traffic understanding.

Traditional machine learning methods~\cite{hayes2016k, van2020flowprint, taylor2017robust, barradas2021flowlens, zhou2023efficient}, constrained by limited model parameters and fitting capabilities, commonly resort to utilizing compressed statistical features at the packet or flow level, such as distributions of packet sizes or inter-arrival times. However, these features often suffer from excessive compression, resulting in the loss of vital information inherent in raw datagrams. 

Recent advancements in deep learning have endeavored to utilize raw traffic bytes. However, as shown in~\cref{tb:rep-comp}, these methods face limitations. They often neglect crucial information in packet headers and introduce unwanted biases by ignoring byte balance or using improper data-splitting techniques.

To address these issues, we propose a novel network traffic representation scheme. 
Our approach remedies the aforementioned shortcomings, preserving hierarchical traffic information while effectively eliminating biases.  

\begin{table}[ht]
    \footnotesize
    \centering
    \caption{Comparison of Existing Representation Schemes}
    \begin{threeparttable}
    \begin{tabular}{ccccc}
        \toprule
         Method & Header & Payload & Byte Balance\tnote{2} & Splitting \\
        \midrule
         PERT\cite{he2020pert} & \ding{55} & \ding{51} & \ding{55} & token \\
         ET-BERT\cite{lin2022bert} & \ding{55} & \ding{51} & \ding{55} & token \\
         YaTC\cite{zhao2023yet} & \ding{51} & \ding{51} & \ding{51} & patch \\
         FlowMAE\cite{hang2023flow} & \ding{51} & \ding{51} & \ding{55} & patch \\
         NetGPT\cite{meng2023netgpt} & \ding{51} & \ding{51} & \ding{55} & token \\
         Lens\cite{wang2024lens} & \ding{51} & \ding{51} & \ding{55} & token \\
         \midrule
         \textbf{\name{}} & \ding{51} & \ding{51} & \ding{51} & stride \\
        \bottomrule
    \end{tabular}
    \begin{tablenotes}
        % \item[1] IA: IP Anonymizing removes all IP addresses.
        \item[1] Byte Balance sets fixed sizes for headers and payloads.
    \end{tablenotes}
    \end{threeparttable}
    \label{tb:rep-comp}
\end{table}
\section{Preliminaries}
This section elaborates on basic definitions, terminologies, and components underlining the Mamba block which serves as the foundation of the proposed \name{}. 
\subsection{State Space Models}
As the key components of Mamba, State Space Models~(SSMs) represent a contemporary category of sequence models within deep learning that share broad connections with Recurrent Neural Networks~(RNNs) and Convolutional Neural Networks~(CNNs). Drawing inspiration from continuous systems, SSMs are commonly structured as linear Ordinary Differential Equations~(ODEs) which establish a mapping from an input sequence $x(t) \in \mathbb{R}^{N}$ to an output sequence $y(t) \in \mathbb{R}^{N}$ via an intermediate latent state $h(t) \in \mathbb{R}^{N}$:

\begin{equation}
    \label{eq:ssm-continuous}
    \begin{aligned}
        h^{'}(t) & =\mathbf{A}h(t) + \mathbf{B}x(t) \\
        y(t) & = \mathbf{C}h(t)
    \end{aligned}
\end{equation}
where $\mathbf{A} \in \mathbb{R}^{N\times N}$ represents the evolution parameter, while $\mathbf{B} \in \mathbb{R}^{N\times 1}$ and $\mathbf{C} \in \mathbb{R}^{1\times N}$ are the projection parameters.

\subsection{Discretization}
Integrating raw SSMs with deep learning presents a significant challenge due to the discrete nature of typical real-world data, contrasting with the continuous-time characteristic of SSMs. To overcome this challenge, the zero-order hold (ZOH) technique is utilized for discretization, leading to the discrete version formulated as follows:

\begin{equation}
\label{eq:ssm-recurrent}
\begin{aligned}
h_{t} & = \mathbf{\overline{A}}h_{t-1} + \mathbf{\overline{B}}x_{t} \\
y_{t} & = \mathbf{C}h_{t}
\end{aligned}
\end{equation}
where $\mathbf{\overline{A}} = \exp(\Delta \mathbf{A})$ and 
% $\mathbf{\overline{B}} = (\Delta \mathbf{A})^{-1}(\exp(\Delta \mathbf{A})-\mathbf{I})\cdot \Delta \mathbf{B} \approx (\Delta \mathbf{A})^{-1}(\Delta \mathbf{A})\Delta \mathbf{B}=\Delta \mathbf{B}$ 
$\mathbf{\overline{B}} \approx \Delta \mathbf{B}$ 
represent the discretized parameters, with $\Delta$ denoting the discretization step size. 
% This recurrent formulation is characterized by linear time complexity.
This recurrent formulation, known for its linear time complexity, is suitable for model inference but lacks parallelizability during training.

By expanding \cref{eq:ssm-recurrent}, SSMs can be transformed into convolutional formulations as follows:
\begin{equation}
\label{eq:ssm-convolutional}
\begin{aligned}
\mathbf{\overline{K}} & = (\mathbf{C}\mathbf{\overline{B}}, \mathbf{C}\mathbf{\overline{A}}\mathbf{\overline{B}}, \dots, \mathbf{C}\mathbf{\overline{A}}^{L-1}\mathbf{\overline{B}}) \\
y & = x * \mathbf{\overline{K}}
\end{aligned}
\end{equation}
where $L$ represents the length of the input sequence $x$,  $*$ denotes the convolution operation, and $\mathbf{\overline{K}} \in \mathbb{R}^{L}$ refers to a structured convolutional kernel. This convolutional representation solves the computational parallelization dilemma encountered in the recurrent version.

\subsection{Selection Mechanism}
 While designed for sequence modeling, SSMs exhibit subpar performance when content-aware reasoning is required, primarily due to their time-invariant nature. Specifically, the parameters $\mathbf{\overline{A}}$, $\mathbf{\overline{B}}$, and $\mathbf{C}$ remain constant across all input tokens within a sequence. To address this issue, Mamba~\cite{gu2023mamba} introduces the selection mechanism, enabling the model to select pertinent information from the context dynamically. This adaptation involves transforming the SSM parameters $\mathbf{\overline{B}}$, $\mathbf{C}$, and $\Delta$ into functions of the input $x$. Additionally, a GPU-friendly implementation is devised to facilitate efficient computation of the selection mechanism, leading to a notable reduction in memory I/O operations and eliminating the need to store intermediate states.

\section{\name{} Framework}

\begin{figure*}[tp]
\centering
\includegraphics[scale=0.6]{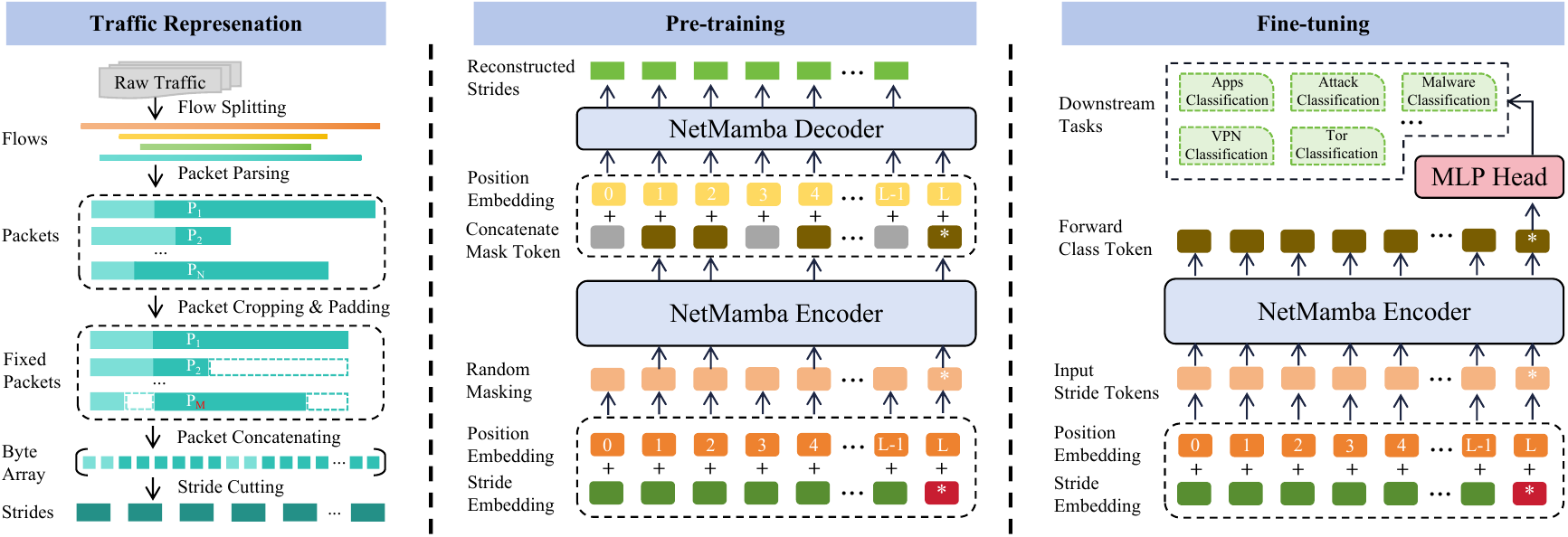}
\caption{Overview of \name{} Framework}
\label{fig:system-design}
\end{figure*}

This section overviews the framework of \name{}~(see \cref{fig:system-design}), providing a comprehensive blueprint for the detailed design presented in \cref{sec:traffic-representation} and \cref{sec:detailed-design}. Initially, \name{} extracts hierarchical information from raw binary traffic and converts it into stride-based representation. Inspired by the Masked AutoEncoders~(MAE) pre-training model in computer vision, \name{} employs a dual-stage training approach. Specifically, self-supervised pre-training is utilized to acquire traffic representation, while supervised fine-tuning is employed to tailor the model for downstream traffic understanding tasks.

\subsubsection{Traffic Representation Phase}
To enhance domain knowledge within networks, \name{} adopts a stride-based methodology to represent key content within network traffic. 
Initially, traffic data is segmented into distinct flows, categorized by their 5-tuple attributes: Source IP, Destination IP, Source Port, Destination Port, and Protocol. 
Fixed-sized segments of header and payload bytes are then extracted for each packet within a flow. 
To collect more comprehensive traffic information without compromising model efficiency due to excessively long packet sequences, we follow approaches outlined in prior studies\cite{lin2022bert, zhao2023yet}, which involve selectively utilizing specific packets within a flow. 
Specifically, bytes from the initial packets of each flow are aggregated into a unified byte array, integrating information across byte, packet, and flow levels for a comprehensive view of traffic characteristics.

This byte array forms the foundation for segmenting non-overlapping flow strides. It preserves semantic relationships between adjacent bytes, effectively mitigating biases introduced by conventional patch-splitting methods, as well as addressing out-of-vocabulary issues commonly associated with tokenization processes. Further design intricacies regarding traffic representation are elucidated in \cref{sec:traffic-representation}.

\subsubsection{Pre-training Phase}
To acquire generic encodings of network domain knowledge based on flow stride representations, \name{} undergoes pre-training using extensive unlabeled network traffic data.
Specifically, \name{} utilizes a masked autoencoder~(MAE) architecture, incorporating multiple unidirectional Mamba blocks in both its encoder and decoder, as detailed in \cref{sec:mamba-block}. 
% Following the original MAE design, \name{} features an asymmetric structure with a heavyweight encoder and a lightweight decoder,
% This configuration is chosen to cultivate a more efficient traffic representation encoder.

% During the pre-training phase, flow strides are processed through several sequential steps. 
% Initially, they are concatenated with a trailing class token and then mapped into stride embeddings. 
% This is followed by the addition of positional embeddings and the random masking of strides before their introduction to the \name{} encoder. 
During pre-training, flow strides undergo several sequential steps: concatenation with a trailing class token, mapping into stride embeddings, addition of positional embeddings, and random masking. 
The encoder focuses solely on visible strides, grasping inherent relationships and generating an output traffic representation. 
% Conversely, the decoder endeavors to reconstruct the masked strides using the output from the encoder in conjunction with dummy masked tokens. 
% The optimization of the pre-training phase is achieved by minimizing the reconstruction loss, particularly for the masked strides, thereby ensuring that the model captures robust representations of traffic patterns. 
The decoder then reconstructs the masked strides using the encoder's output and dummy tokens. Pre-training is optimized by minimizing the reconstruction loss for the masked strides, ensuring the model learns robust traffic patterns.
Detailed insights into the pre-training strategy are provided in \cref{sec:flowmamba-pretraining}.

\subsubsection{Fine-tuning Phase}
For accurately capturing traffic patterns and understanding downstream task requirements, \name{} undergoes fine-tuning using labeled traffic data.
During this phase, the decoder of \name{} is replaced by a multi-layer perceptron~(MLP) head to facilitate classification tasks. 
With the removal of the reconstruction task, all embedded flow strides become visible to the encoder. 
As the unidirectional Mamba block processes sequence information in a front-to-back manner, the trailing class token, after being processed by the encoder, aggregates the overall traffic characteristics.
Subsequently, \name{} forwards only this class token to the MLP-based classifier.

% Following pre-training, the encoder of \name{} demonstrates significant adaptability when fine-tuned with a limited quantity of labeled, task-specific traffic data. This adaptability allows \name{} to efficiently transition to a variety of downstream tasks, including but not limited to application classification and attack detection. For additional information concerning the fine-tuning process, please refer to Section \ref{sec:flowmamba-finetuning}.
Post pre-training, \name{}'s encoder exhibits significant adaptability when fine-tuned with limited labeled data, enabling efficient transition to various downstream tasks such as application classification and attack detection. Additional details on the fine-tuning process are provided in \cref{sec:flowmamba-finetuning}.
\section{Traffic Representation \label{sec:traffic-representation}}
This section provides detailed information about the traffic representation scheme used by \name{}. The key hyper-parameters are listed in \cref{tb:notation}.
% This approach retains header information, mitigates IP biases, ensures balanced byte allocation, and tailors data-splitting techniques. Ultimately, this method preserves hierarchical traffic information while eliminating biases effectively. 

\begin{table}[ht]
    \footnotesize
    \centering
    \caption{Summary of Hyper-Parameter Notations in \name{}}
    \begin{threeparttable}
    \begin{tabular}{c|c}
        \toprule
         Notation & Description \\
        \midrule
         $M$ & Number of packets selected from a single flow \\ \midrule
         $N_h$ & Number of header bytes selected from a single packet \\ \midrule
         $N_p$ & Number of payload bytes selected from a single packet \\ \midrule
         $L_b$ & Length of the byte array for a single flow \\ \midrule
         $L_s$ & Length of the consecutive bytes for a flow stride \\ \midrule
         $N_s$ & Length of the stride sequence for a single flow \\ \midrule
         $\mathtt{D}_{\text{enc}} / \mathtt{D}_{\text{dec}} $ & Hidden state dimension of \name{} encoder or decoder  \\ \midrule
         $\mathtt{E}_{\text{enc}} / \mathtt{E}_{\text{dec}} $ & Expanded state dimension of \name{} encoder or decoder \\ \midrule
         $\mathtt{N}$ & Dimension of state space models in \name{}  \\ \midrule
         $\mathtt{B}$ & Batch size of the input token sequence  \\ \midrule
         $\mathtt{L}$ & Length of the original input token sequence  \\ \midrule
         $r$ & Ratio of masked stride tokens  \\ \midrule
         $\mathtt{L}_{\text{vis}}$ & Length of the visible input token sequence  \\
        \bottomrule
    \end{tabular}
    % \begin{tablenotes}
    % \end{tablenotes}
    \end{threeparttable}
    \label{tb:notation}
\end{table}

\subsubsection{Flow Splitting}
Formally, given network traffic comprising multiple packets, we segment it into various flows, with each flow consisting of packets that belong to a specific protocol and are transmitted between two ports on two hosts. Packets within the same flow encapsulate significant interaction information between the two hosts. This information includes the establishment of a TCP connection, data exchanged during communication, and the overall transmission status. These flow-level features are pivotal in characterizing application behaviors and enhancing the efficiency of traffic classification processes.

\subsubsection{Packet Parsing}
For each flow, all packets are processed through several sequential operations to preserve valuable information and eliminate unnecessary interference. 
When narrowing down the scope for analyzing traffic data related to specific applications or services, we exclude all packets carried by non-IP protocols, such as Address Resolution Protocol~(ARP) and Dynamic Host Configuration Protocol~(DHCP). 
Considering the critical information contained within both the packet (e.g., the total length field) and the payload (text content for upper-level protocols), we choose to retain these elements.
Furthermore, to mitigate biases introduced by identifiable information, all packets are anonymized through the removal of 
% potential 
Ethernet headers. 
% and the masking of IP addresses.

\subsubsection{Packet Cropping \& Padding, and Concatenating}
Given the variability in packet size within the same flow and the fluctuation in both header length (including the IP header and any 
% potential 
upper-layer headers) and payload length within individual packets, problematic scenarios often arise. For instance, the first long packet can occupy the entire limited model input array, or excessively long payloads can dominate the byte information within shorter headers. Therefore, it is essential to standardize packet sizes by assigning uniform sizes to all packets and fixed lengths to both packet headers and payloads.
Specifically, we select the first $M$ packets from a single flow, setting the header length to $N_h$ bytes and the payload length to $N_p$ bytes. 
Any packet exceeding this length will be cropped, while shorter packets will be padded to meet these specifications. 

Eventually, all bytes of initial $M$ packets are concatenated into an unified array $[b_1, b_2, \dots, b_{L_b}]$ where $L_b = M \times (N_h + N_p)$ represents the array length and $b_i$ denotes the $i$-th byte. 

\subsubsection{Stride Cutting}
Given the significant computational and memory demands posed by a byte array with $L_b$ (typically greater than 1000) elements, it becomes imperative to explore further compression techniques to enhance the efficiency of model training and inference. Traditional methods often involve reshaping the byte array into a square matrix and employing two-dimensional patch splitting, a practice borrowed from computer vision. However, this technique unintentionally introduces biases by grouping vertically adjacent bytes that are semantically unrelated, as they are not naturally contiguous in the sequential traffic data.
% However, this approach inadvertently introduces biases by grouping vertically adjacent bytes that are semantically unrelated.

Inspired by patching methods used in time-series forecasting, we adopt a 1-dimensional stride cutting approach on the original array, aligning with the sequential nature of network traffic and preserving inter-byte correlations. 
Specifically, we divide the byte array into non-overlapping strides of size $1 \times L_s$, resulting in a total number of strides $N_s = L_b / L_s$. Each stride $\mathbf{s}_i \in \mathbb{R}^{1 \times L_s}$ is defined as $[b_{L_s \times i}, b_{L_s \times i + 1}, \dots, b_{L_s \times (i+1) - 1}]$ for $0 \le i < N_s$. 
This strategy aims to mitigate biases while retaining essential sequential information in the data.

\textbf{Takeaway}. \emph{Our traffic representation scheme effectively retains crucial information from both packet headers and payloads, while eliminating unwanted biases through techniques such as IP anonymization, byte balancing, and stride cutting. For a detailed evaluation, please refer to \cref{sec:ab-study}.}
\section{Model Details \label{sec:detailed-design}}
This section details the \name{} model architecture, along with the pre-training and fine-tuning strategies.

\subsection{\name{} Architecture}
\begin{figure}[tp]
\centering
\includegraphics[scale=0.45]{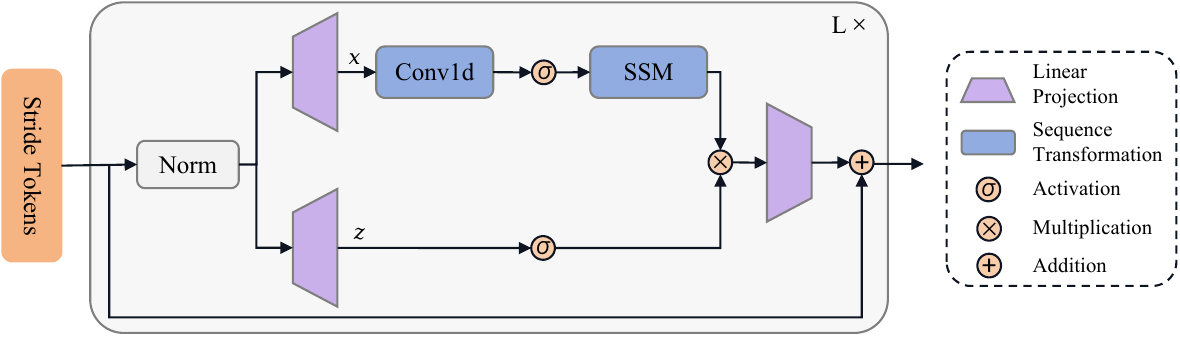}
\caption{\name{} Encoder (Decoder)}
\label{fig:traffic-mamba}
\end{figure}

\subsubsection{Stride Embedding}
Given the stride array, we initially perform a linear projection on  each stride $\mathbf{s}_i$ to a vector with size $\mathtt{D}_{\text{enc}}$ and incorporate position embeddings $\mathbf{E}_{\text{enc}}^{\text{pos}} \in \mathbb{R}^{N_s  \times \mathtt{D}_{\text{enc}}}$ as shown below:
\begin{equation}
    \mathbf{X}_0 = \left[ \mathbf{s}_1\mathbf{W}; \mathbf{s}_2\mathbf{W}; \cdots; \mathbf{s}_{N_s}\mathbf{W}; \mathbf{x}_{\text{cls}}\right] + \mathbf{E}_{\text{enc}}^{\text{pos}}
\end{equation}
where $\mathbf{W} \in \mathbb{R}^{L_s \times \mathtt{D}_{\text{enc}}}$ represents the learnable projection matrix. Inspired by ViT~\cite{dosovitskiy2020image} and BERT~\cite{devlin2018bert}, we introduce a class token to represent the entire stride sequence, denoted as $\mathbf{x}_{\text{cls}}$. Since the unidirectional Mamba processes sequence information from front to back, we opt to append the class token to the end of the sequence for enhanced information aggregation.

\subsubsection{\name{} Block \label{sec:mamba-block}}
Recently, several variants of Mamba have been proposed to accommodate domain-specific data formats and task requirements. For instance, Vim~\cite{zhu2024vision} incorporates bidirectional Mamba blocks for spatial-aware understanding of vision tasks, Graph-Mamba~\cite{wang2024graph} introduces a graph-dependent selection mechanism for graph learning, while MiM-ISTD~\cite{chen2024mim} customizes a cascading Mamba structure for extracting hierarchical visual information. We argue that the original unidirectional Mamba design\cite{gu2023mamba}, tailored for sequence modeling, is well-suited for representation learning in sequential network traffic, offering increased efficiency through the elimination of omnidirectional scans and redundant blocks. We carefully test different Mamba variants, demonstrating that the selected unidirectional Mamba is more suitable for processing network traffic. Please refer to the ablation studies for more details.

Hence, we implement the \name{} encoder and decoder using unidirectional Mamba blocks, as illustrated in \cref{fig:traffic-mamba}. The operational process of the \name{} block forward pass is outlined in \cref{alg:block}. For a given input token sequence $\mathbf{X}_{t-1}$ with a batch size $\mathtt{B}$ and sequence length $\mathtt{L}$ from the $(t-1)$-th \name{} block, we begin by normalizing it and then projecting it linearly into $\mathbf{x}$ and $\mathbf{z}$, both with dimension size of $\mathtt{E}$. We subsequently apply causal 1-D convolution to $\mathbf{x}$, resulting in $\mathbf{x}'$. Based on $\mathbf{x}'$, we compute the input-dependent step size $\mathbf{\Delta}$, as well as the projection parameters $\mathbf{B}$ and $\mathbf{C}$ having a dimension size of 
 $\mathtt{N}$. We then discretize $\overline{\mathbf{A}}$ and $\overline{\mathbf{B}}$ using $\mathbf{\Delta}$. Following this, we calculate $\mathbf{y}$ employing a hardware-aware SSM. Finally, $\mathbf{y}$ is gated by $\mathbf{z}$ and added residually to $\mathbf{X}_{t-1}$, resulting in the output token sequence $\mathbf{X}_{t}$ for the $t$-th \name{} block. 

\begin{algorithm}[ht]
\caption{\name{} Block Forward Pass}
\small
\begin{algorithmic}[1]
\REQUIRE{$\mathbf{X}_{t-1}$ : \textcolor{shapecolor}{$(\mathtt{B}, \mathtt{L}, \mathtt{D})$}}
\ENSURE{$\mathbf{X}_{t}$ : \textcolor{shapecolor}{$(\mathtt{B}, \mathtt{L}, \mathtt{D})$}}
\STATE $\mathbf{X}_{t-1}'$ : \textcolor{shapecolor}{$(\mathtt{B}, \mathtt{L}, \mathtt{D})$} $\leftarrow$ $\mathbf{Norm}(\mathbf{X}_{t-1})$ \COMMENT{normalize input sequence}
\STATE $\mathbf{x}$ : \textcolor{shapecolor}{$(\mathtt{B}, \mathtt{L}, \mathtt{E})$} $\leftarrow$ $\mathbf{Linear}^\mathbf{x}(\mathbf{X}_{t-1}')$ 
\STATE $\mathbf{z}$ : \textcolor{shapecolor}{$(\mathtt{B}, \mathtt{L}, \mathtt{E})$} $\leftarrow$ $\mathbf{Linear}^\mathbf{z}(\mathbf{X}_{t-1}')$
\STATE $\mathbf{x}'$ : \textcolor{shapecolor}{$(\mathtt{B}, \mathtt{L}, \mathtt{E})$} $\leftarrow$ $\mathbf{SiLU}(\mathbf{Conv1d}(\mathbf{x}))$
\STATE $\mathbf{B}$ : \textcolor{shapecolor}{$(\mathtt{B}, \mathtt{L}, \mathtt{N})$} $\leftarrow$ $\mathbf{Linear}^{\mathbf{B}}(\mathbf{x}')$ \COMMENT{input-dependent}
\STATE $\mathbf{C}$ : \textcolor{shapecolor}{$(\mathtt{B}, \mathtt{L}, \mathtt{N})$} $\leftarrow$ $\mathbf{Linear}^{\mathbf{C}}(\mathbf{x}')$ \COMMENT{input-dependent}
\STATE $\mathbf{\Delta}$ : \textcolor{shapecolor}{$(\mathtt{B}, \mathtt{L}, \mathtt{E})$} $\leftarrow$ $\log(1 + \exp(\mathbf{Linear}^{\mathbf{\Delta}}(\mathbf{x}') + \mathbf{Parameter}^{\mathbf{\Delta}}))$ \COMMENT{softplus ensures positive step size, input-dependent}
\STATE $\overline{\mathbf{A}}$ : \textcolor{shapecolor}{$(\mathtt{B}, \mathtt{L}, \mathtt{E}, \mathtt{N})$} $\leftarrow$ $\mathbf{\Delta} \bigotimes \mathbf{Parameter}^{\mathbf{A}}$ \COMMENT{discritize}
\STATE $\overline{\mathbf{B}}$ : \textcolor{shapecolor}{$(\mathtt{B}, \mathtt{L}, \mathtt{E}, \mathtt{N})$} $\leftarrow$ $\mathbf{\Delta} \bigotimes \mathbf{B}$ \COMMENT{discritize}
\STATE $\mathbf{y}$ : \textcolor{shapecolor}{$(\mathtt{B}, \mathtt{L}, \mathtt{E})$} $\leftarrow$ $\mathbf{SSM}(\overline{\mathbf{A}}, \overline{\mathbf{B}}, \mathbf{C})(\mathbf{x}')$ \COMMENT{hardware-aware scan}
\STATE $\mathbf{y}'$ : \textcolor{shapecolor}{$(\mathtt{B}, \mathtt{L}, \mathtt{E})$} $\leftarrow$ $\mathbf{y} \bigodot \mathbf{SiLU}(\mathbf{z}) $ \COMMENT{self-gating}
\STATE $\mathbf{X}_{t}$ : \textcolor{shapecolor}{$(\mathtt{B}, \mathtt{L}, \mathtt{D})$} $\leftarrow$ $\mathbf{Linear}^\mathbf{X}(\mathbf{y}') + \mathbf{X}_{t-1}$ \COMMENT{residual connection}
\STATE Return: $\mathbf{X}_{t}$ \COMMENT{output sequence}
\end{algorithmic}
\label{alg:block}
\end{algorithm}

% \subsection{Training Strategy}
\subsection{\name{} Pre-training \label{sec:flowmamba-pretraining}}
\subsubsection{Random Masking}
Given the embedded stride tokens $\mathbf{X}_0 \in \mathbb{R}^{\mathtt{L}\times \mathtt{D}_{\text{enc}}}$, a portion of strides is randomly sampled while the remaining ones are removed. For a predefined masking ratio $r \in (0, 1)$, the length of visible tokens is determined as $\mathtt{L}_{\text{vis}} = \lceil(1-r)\mathtt{L}\rceil$. The visible tokens are then sampled as follows:
\begin{equation}
    \mathbf{X}_0^{\text{vis}} = \mathbf{Shuffle}(\mathbf{X}_0)[1:\mathtt{L}_{\text{vis}}, \ :\ ]
    \in \mathbb{R}^{\mathtt{L}_{\text{vis}}\times \mathtt{D}_{\text{enc}}}
\end{equation}
where the $\mathbf{Shuffle}$ operation permutes the token sequence randomly. Notably, we ensure that the trailing class token remains unmasked throughout this process since its role in aggregating overall sequence information necessitates its preservation at all times.

The primary objective behind random masking is the elimination of redundancy. This approach creates a challenging task that resists straightforward solutions through extrapolation from neighboring strides alone. 
Additionally, the reduction in input length diminishes computational and memory costs, offering an opportunity for more efficient model training. 

\subsubsection{Masked Pre-training}
The \name{} encoder is tasked with capturing latent inter-stride relationships using the visible tokens, whereas the \name{} decoder's objective is to reconstruct masked strides utilizing both the encoder output tokens and mask tokens. Each mask token represents a shared, trainable vector indicating the presence of a missing stride. Additionally, new positional embeddings are added to provide location information to the mask tokens.

The formal forward process of \name{} pre-training can be outlined as follows:
\begin{equation}
\begin{aligned}
    \mathbf{X}_{\text{enc}}^{\text{out}} & = \mathbf{MLP}(\mathbf{Encoder}(\mathbf{X}_0^{\text{vis}})) \in \mathbb{R}^{\mathtt{L}_{\text{vis}}\times \mathtt{D}_{\text{dec}}}\\
    \mathbf{X}_{\text{dec}}^{\text{in}} & = \mathbf{Unshuffle}(\mathbf{Concat}(\mathbf{X}_{\text{enc}}^{\text{out}}, \mathbf{X}_{\text{mask}})) + \mathbf{E}_{\text{dec}}^{\text{pos}} \\
    \mathbf{X}_{\text{dec}}^{\text{out}} & = \mathbf{Decoder}(\mathbf{X}_{\text{dec}}^{\text{in}})
\end{aligned}
\end{equation}
where the $\mathbf{Unshuffle}$ operation restores the original sequence order, and $\mathbf{E}_{\text{dec}}^{\text{pos}} \in \mathbb{R}^{\mathtt{L}\times \mathtt{D}_{\text{dec}}}$ represents decoder-specific positional embeddings. Subsequently, the mean square error~(MSE) loss for self-supervised reconstruction is calculated as shown below:
\begin{equation}
\begin{aligned}
    \mathbf{y}_{\text{real}} & = \mathbf{Shuffle}(\mathbf{X}_0)[\mathtt{L}_{\text{vis}}+1:\mathtt{L}, \ :\ ] \\
    \mathbf{y}_{\text{rec}} & = \mathbf{Shuffle}(\mathbf{X}_{\text{dec}}^{\text{out}})[\mathtt{L}_{\text{vis}}+1:\mathtt{L}, \ :\ ] \\
    \mathcal{L}_{\text{rec}} & = \mathbf{MSE}(\mathbf{y}_{\text{real}}, \mathbf{y}_{\text{rec}})
\end{aligned}
\end{equation}
where $\mathbf{y}_{\text{real}}$ represents the ground-truth mask tokens, and $\mathbf{y}_{\text{rec}}$ signifies the predicted ones.

\subsection{\name{} Fine-tuning \label{sec:flowmamba-finetuning}}
For downstream tasks, all encoder parameters, including embedding modules and Mamba blocks, are loaded from pre-training. To conduct classification on labeled traffic data, the decoder is replaced with an MLP head. Given that all stride tokens are visible, fine-tuning of \name{} is performed in a supervised manner as detailed below:
\begin{equation}
\begin{aligned}
    \mathbf{X} & = \mathbf{Encoder}(\mathbf{X}_0) \in \mathbb{R}^{\mathtt{L}\times \mathtt{D}_{\text{enc}}} \\
    % \mathbf{f} & = \mathbf{X}[\mathtt{L}, \ :] \in \mathbb{R}^{\mathtt{D}_{\text{enc}}} \\
    % \hat{\mathbf{y}} & = \mathbf{MLP(\mathbf{Norm}(\mathbf{f}))}
    \hat{\mathbf{y}} & = \mathbf{MLP(\mathbf{Norm}(\mathbf{X}[\mathtt{L}, \ :]))}
\end{aligned}
\end{equation}
Here, $\mathbf{f}$ denotes the trailing class token, and $\hat{\mathbf{y}} \in \mathbb{R}^{\mathtt{C}}$ represent the prediction distribution, where $\mathtt{C}$ is the number of traffic categories. The classification process is then optimized by minimizing the cross-entropy loss between the prediction distribution $\hat{\mathbf{y}}$ and the ground-truth label $\mathbf{y}$:
\begin{equation}
\mathcal{L}_{\text{cls}} = \mathbf{CrossEntropy}(\hat{\mathbf{y}}, \mathbf{y})
\end{equation}

\textbf{Takeaway}. \emph{The unidirectional Mamba architecture is well-suited for processing sequential network traffic data. To acquire generic network domain knowledge, \name{} is pre-trained by reconstructing masked strides. For adaptation to specific downstream tasks, \name{} is fine-tuned by minimizing prediction loss.}
\section{Evaluation}
\subsection{Experimental Setup}
\subsubsection{Datasets}
To assess the effectiveness and generalization abilities of \name{}, we conducted experiments using six publicly available real-world traffic datasets encompassing three main
classification tasks.
\begin{enumerate}[label=\arabic*.]
    \item \textbf{Encrypted Application Classification}: This task aims to classify application traffic under various encryption protocols. Specifically, the CrossPlatform (Android) \cite{ren2019international} and CrossPlatform (iOS) \cite{ren2019international} contain 254 and 253 applications respectively. Additionally, we use Tor traffic data from 8 communication categories in ISCXTor2016 \cite{lashkari2017characterization} and VPN traffic data from 7 communication categories in ISCXVPN2016 \cite{gil2016characterization}. 
    \item \textbf{Attack Traffic Classification}: This task aims to identify potential attack traffic, such as Denial of Service~(DoS) attacks and brute force attacks. We construct 6 data categories using CICIoT2022\cite{dadkhah2022towards}.
    \item \textbf{Malware Traffic Classification}: This task aims to distinguish between traffic generated by malware and benign traffic. We use all 20 data categories from the USTC-TFC2016 dataset \cite{wang2017malware}.
\end{enumerate}
We observed an imbalance in flow counts across traffic categories, which adversely impacts model performance. To address this, we set upper and lower flow limits for each category. Categories below the lower limit are discarded, while those above the upper limit are randomly sampled.
% We have noted a significant discrepancy in the number of flows across categories within a given dataset, which adversely affects the performance and robustness of a traffic understanding model. To address the data distribution imbalance,  we set lower and upper limits for each category within the dataset. Any traffic category with a flow amount below the lower limit is discarded, while those exceeding one are randomly sampled to meet the criteria. 
% The statistical details of our normalized datasets are presented in \cref{tb:datasets}. 
% \begin{table}[ht]
%     \footnotesize
%     \centering
%     \caption{The Statistical Information of Normalized Datasets}
%     \begin{tabular}{ccccc}
%         \toprule
%          Dataset & Lower & Upper & \# Flow & \# Category \\
%         \midrule
%          CrossPlatform(Android) & 50 & 2,000 & 17,624 & 181 \\
%          CrossPlatform(iOS) & 50 & 2,000 & 9,536 & 124 \\
%          CICIoT2022 & 200 & 6,000 & 19,497 & 6 \\
%          ISCXTor2016 & 10 & 4,000 & 188 & 8 \\
%          ISCXVPN2016 & 500 & 4,000 & 13,832 & 7 \\
%          USTC-TFC2016 & 500 & 2,000 & 40,000 & 20 \\
%         \bottomrule
%     \end{tabular}
%     \label{tb:datasets}
% \end{table}

\subsubsection{Comparison Methods}
To comprehensively evaluate \name{}, we conducted comparisons with various open-source baselines and state-of-the-art techniques, as outlined below:
\begin{enumerate}[label=\arabic*.]
    \item Classical machine learning methods such as \textbf{AppScanner} \cite{taylor2017robust} and \textbf{FlowPrint} \cite{van2020flowprint} that rely on statistical features for traffic classification.
    \item Deep learning approaches like \textbf{FS-Net} \cite{liu2019fs}
    % , \textbf{DeepPacket}\cite{lotfollahi2020deep} 
    and \textbf{TFE-GNN} \cite{zhang2023tfe} that utilize packet lengths or raw bytes to perform traffic analysis in a supervised manner.
    \item Transformer-based models such as \textbf{ET-BERT} \cite{lin2022bert} and \textbf{YaTC} \cite{zhao2023yet} that capture traffic representations during pre-training and subsequently fine-tune for specific tasks with limited labeled data. In particular, we implement \textbf{YaTC(OF)} by substituting packet-level and flow-level attention with a global attention module, which expedites model inference while removing its original memory optimization.
    \item Transformer variants within the \name{} backbone, including \textbf{NT-Vanilla} and \textbf{NT-Linear}. The former replaces Mamba blocks in \name{} with vanilla Transformer blocks \cite{vaswani2017attention} featuring quadratic complexity, while the latter adopts Linear Transformer blocks \cite{katharopoulos2020transformers} with linear complexity.
\end{enumerate}

\subsubsection{Implementation Details}
At the pre-training stage, we set the batch size to $\mathtt{B} = 128$ and train models for 150,000 steps. The initial learning rate is set to $1.0 \times 10^{-3}$ with the AdamW optimizer, alongside a linear learning rate scaling policy. Additionally, a masking ratio of $r = 0.9$ is employed for randomly masking strides.

For fine-tuning, we adjust the batch size to $\mathtt{B} = 64$ and set the learning rate to $2.0 \times 10^{-3}$. Each dataset is partitioned into training, validation, and test sets following an 8:1:1 ratio. All models are trained for 120 epochs on the training data, with checkpoints saving the best accuracy on the validation set, subsequently evaluated on the test set.

The \name{} architecture features an encoder composed of 4 Mamba blocks and a decoder composed of 2 Mamba blocks. More hyper-parameter details can be found in \cref{tb:hyper-param}.

The proposed model is implemented using PyTorch 2.1.1, with all experiments conducted on a Ubuntu 22.04 server equipped with CPU of Intel(R) Xeon(R) Gold 6240C CPU @ 2.60GHz, GPU of NVIDIA A100 (40GB $\times$ 4).

\begingroup
\renewcommand{\arraystretch}{1.3}
\begin{table}[ht]
    \footnotesize
    \centering
    \caption{Hyper-Parameter details of \name{}}
    \begin{tabular}{c|c||c|c||c|c}
        \toprule
        \hline
         Variable & Value & Variable & Value & Variable & Value \\ \hline
         $M$ & 5 & $\mathtt{D}_{\text{enc}}$ & 256 & $L_s$ & 4 \\ \hline
         $N_h$ & 80 & $\mathtt{D}_{\text{dec}}$ & 128 & $\mathtt{N}$ & 16 \\ \hline
         $N_p$ & 240 & $\mathtt{E}_{\text{enc}}$ & 512 & $\mathtt{L}$ & 401 \\ \hline
         $L_b$ & 1600 & $\mathtt{E}_{\text{dec}}$ & 256  & $\mathtt{L}_{\text{vis}}$ & 41 \\ \hline
        \bottomrule
    \end{tabular}
    \label{tb:hyper-param}
\end{table}
\endgroup

\subsubsection{Evaluation Metrics}
We assess the performance of \name{} using four typical metrics: Accuracy(AC), Precision(PR), Recall(RC), and weighted F1 Score(F1).

\subsection{Overall Evaluation}
\begin{table*}[h]
  \centering
  \caption{Comparison Results on CrossPlatform(Android), CrossPlatform(iOS) and CICIoT2022}
  \begin{tabular}{c|cc|cccc|cccc|cccc}
    \toprule
    \multirow{2}{*}{Method} & \multicolumn{2}{c|}{Params(M)} & \multicolumn{4}{c|}{CrossPlatform(Android)\cite{ren2019international}} & \multicolumn{4}{c|}{CrossPlatform(iOS)\cite{ren2019international}} & \multicolumn{4}{c}{CICIoT2022 \cite{dadkhah2022towards}} \\
    \cline{2-15} \noalign{\vskip 0.5mm}
     & PT & FT & AC & PR & RC & F1 & AC & PR & RC & F1 & AC & PR & RC & F1 \\
     \midrule
     AppScanner\cite{taylor2017robust} & - & - & 0.1626 & 0.1646 & 0.1456 & 0.1413 & 0.1718 & 0.1400 & 0.1440 & 0.1283 & 0.7556 & 0.8093 & 0.7244 & 0.6938 \\
     FlowPrint\cite{van2020flowprint} & - & - & 0.8739 & 0.8941 & 0.8739 & 0.8700 & 0.8712 & 0.8687 & 0.8712 & 0.8603 & 0.5820 & 0.4164 & 0.5820 & 0.4643 \\
     \midrule
     FS-Net\cite{liu2019fs} & - & 5.3 & 0.0147 & 0.0023 & 0.0147 & 0.0034 & 0.0293 & 0.0014 & 0.0293 & 0.0025 & 0.5747 & 0.3800 & 0.5747 & 0.4216 \\
     TFE-GNN\cite{zhang2023tfe} & - & 44.3 & 0.8141 & 0.8308 & 0.8141 & 0.8067 & 0.8241 & 0.8326 & 0.8241 & 0.8130 & \cellcolor{lightgray}1.000 & \cellcolor{lightgray}1.000 & \cellcolor{lightgray}1.000 & \cellcolor{lightgray}1.000 \\
     \midrule
     ET-BERT\cite{lin2022bert} & 187.4 & 136.4 & 0.8743 & 0.8913 & 0.8743 & 0.8786 & 0.9105 & 0.8809 & 0.9105 & 0.8850 & 0.9937 & 0.9938 & 0.9937 & 0.9937 \\
     YaTC(OF)\cite{zhao2023yet} & 2.3 & 2.1 & \underline{0.9076} & \underline{0.9107} & \underline{0.9076} & \underline{0.9077} & 0.9263 & 0.9282 & 0.9263 & 0.9264 & 0.9949 & 0.9949 & 0.9949 & 0.9949 \\
     YaTC \cite{zhao2023yet} & 2.3 & 2.1 & 0.8952 & 0.8989 & 0.8952 & 0.8952 & \underline{0.9270} & \underline{0.9296} & \underline{0.9270} & \underline{0.9272} & \underline{0.9974} & \underline{0.9975} & \underline{0.9974} & \underline{0.9974} \\
     \midrule
     \textbf{\name{}} & 2.2 & 1.9 & \cellcolor{lightgray}0.9094 & \cellcolor{lightgray}0.9133 & \cellcolor{lightgray}0.9094 & \cellcolor{lightgray}0.9096 & \cellcolor{lightgray}0.9301 & \cellcolor{lightgray}0.9327 & \cellcolor{lightgray}0.9301 & \cellcolor{lightgray}0.9305 & 0.9928 & 0.9931 & 0.9928 & 0.9929 \\
    \bottomrule
  \end{tabular}
  \label{tb:overall-1}
\end{table*}

\begin{table*}[h]
  \centering
  \caption{Comparison Results on ISCXTor2016, ISCXVPN2016 and USTC-TFC2016}
  \begin{tabular}{c|cc|cccc|cccc|cccc}
    \toprule
    \multirow{2}{*}{Method} & \multicolumn{2}{c|}{Params(M)} & \multicolumn{4}{c|}{ISCXTor2016 \cite{lashkari2017characterization}} & \multicolumn{4}{c|}{ISCXVPN2016 \cite{gil2016characterization}} & \multicolumn{4}{c}{USTC-TFC2016 \cite{wang2017malware}} \\
    \cline{2-15} \noalign{\vskip 0.5mm}
     & PT & FT & AC & PR & RC & F1 & AC & PR & RC & F1 & AC & PR & RC & F1 \\
     \midrule
     AppScanner\cite{taylor2017robust} & - & - & 0.4034 & 0.2850 & 0.2149 & 0.2113 & 0.7643 & 0.8047 & 0.7045 & 0.7256 & 0.6998 & 0.8591 & 0.6062 & 0.6633 \\
     FlowPrint\cite{van2020flowprint} & - & - & 0.1316 & 0.0173 & 0.1316 & 0.0306 & 0.9666 & 0.9733 & 0.9666 & 0.9681 & 0.7992 & 0.7745 & 0.7992 & 0.7755 \\
     \midrule
     FS-Net\cite{liu2019fs} & - & 5.3 & 0.7020 & 0.7010 & 0.7020 & 0.6999 & 0.7023 & 0.7487 & 0.7023 & 0.6660 & 0.4381 & 0.2011 & 0.4381 & 0.2672 \\
     TFE-GNN\cite{zhang2023tfe} & - & 44.3 & 0.7692 & 0.8030 & 0.7692 & 0.7618 & 0.8428 & 0.8508 & 0.8428 & 0.8447 & 0.9747 & 0.9747 & 0.9747 & 0.9734 \\
     \midrule
     ET-BERT\cite{lin2022bert} & 187.4 & 136.4 & \underline{0.9967} & \underline{0.9967} & \underline{0.9967} & \underline{0.9967} & 0.9566 & 0.9566 & 0.9566 & 0.9565 & 0.9910 & 0.9911 & 0.9910 & 0.9910 \\
     YaTC(OF)\cite{zhao2023yet} & 2.3 & 2.1 & \cellcolor{lightgray}0.9986 & \cellcolor{lightgray}0.9986 & \cellcolor{lightgray}0.9986 & \cellcolor{lightgray}0.9986 & \underline{0.9805} & \underline{0.9808} & \underline{0.9805} & \underline{0.9806} & \underline{0.9960} & 0.9955 & \underline{0.9960} & \underline{0.9957} \\
     YaTC \cite{zhao2023yet} & 2.3 & 2.1 & 0.9959 & 0.9959 & 0.9959 & 0.9959 & \cellcolor{lightgray}0.9848 & \cellcolor{lightgray}0.9849 & \cellcolor{lightgray}0.9848 & \cellcolor{lightgray} \cellcolor{lightgray}0.9848 & \cellcolor{lightgray}0.9972 & \cellcolor{lightgray}0.9976 & \cellcolor{lightgray}0.9972 & \cellcolor{lightgray}0.9970 \\
    \midrule
    \textbf{\name{}} & 2.2 & 1.9 & \cellcolor{lightgray}0.9986 & \cellcolor{lightgray}0.9986 & \cellcolor{lightgray}0.9986 & \cellcolor{lightgray}0.9986 & \underline{0.9805} & \underline{0.9808} & \underline{0.9805} & \underline{0.9806} & \underline{0.9960} & \underline{0.9957} & \underline{0.9960} & \underline{0.9957} \\
    \bottomrule
  \end{tabular}
  \label{tb:overall-2}
\end{table*}

We evaluated the performance of \name{} in categorizing traffic using six publicly available datasets. As shown in \cref{tb:overall-1} and \cref{tb:overall-2}, \name{} consistently outperforms all baseline methods on three datasets and ranks second on two others. However, it falls slightly short on the CICIoT2022 dataset, with a maximum difference of 0.72 percentage points in both accuracy and F1 score. On average, \name{} achieves accuracy levels between 0.9094 and 0.9986, and F1 scores ranging from 0.9096 to 0.9986. Notably, \name{} maintains the fewest parameters among all deep learning methods, underscoring its efficient yet effective capabilities in traffic representation learning.

% \subsubsection{CrossPlatform (Android)} 
% As indicated in \cref{tb:overall-1}, \name{} demonstrates significant improvements over existing methods on the CrossPlatform(Android) dataset. Specifically, our model achieves a notable improvement of 4.83\% in accuracy and 4.64\% in f1 score compared to the state-of-the-art method (ET-BERT). ET-BERT focuses solely on learning traffic representations from packet payloads, overlooking the valuable content information carried by packet headers. In contrast, \name{} effectively models both header and payload characteristics, leading to a more comprehensive analysis of traffic patterns.

% \subsubsection{CrossPlatform (iOS)} 
% On the CrossPlatform(iOS) dataset, \name{} surpasses all baseline methods and outperforms the state-of-the-art technique (YaTC) by more than 5\% across all evaluation metrics. Beyond the variances in base model architecture, \name{} establishes a more resilient traffic representation scheme compared to YaTC. This enhancement is achieved through techniques such as IP masking and stride cutting, contributing to a stronger overall performance.

\subsubsection{CrossPlatform (Android \& iOS) and ISCXTor2016} 
The CrossPlatform Android and iOS datasets consist of encrypted traffic generated by the top 100 applications from the US, China, and India, encompassing over 200 categories in both cases. Additionally, ISCXTor2016 includes application traffic using the Onion Router~(Tor) for encrypted communications.

As shown in \cref{tb:overall-1}, \name{} demonstrates superior performance across these three datasets, with F1 score improvements ranging from 0.18\% to 0.31\%. In comparison to state-of-the-art pre-trained methods, ET-BERT exclusively focuses on learning traffic representations from packet payloads, overlooking the valuable information carried by packet headers. Moreover, beyond the variances in base model architecture, YaTC differs from \name{} in the traffic representation scheme, employing a two-dimensional splitting technique. In contrast, \name{} effectively models both header and payload characteristics and optimizes traffic data segmentation to mitigate biases. This leads to a more comprehensive analysis of traffic patterns.

\subsubsection{CICIoT2022, ISCXVPN2016 \& USTC-TFC2016}
The CICIoT2022 dataset consists of traffic collected from a laboratory network designed for profiling, behavioral analysis, and vulnerability testing of various IoT devices. The ISCXVPN2016 dataset includes encrypted communication traffic tunneled through Virtual Private Networks~(VPN). The USTC-TFC2016 dataset comprises encrypted traffic from both malware and benign applications.

As depicted in \cref{tb:ablation}, \name{} performs only slightly worse than YaTC on the ISCXVPN2016 and USTC-TFC2016 datasets, while achieving comparable performance to other state-of-the-art methods on the CICIoT2022 dataset. Although TFE-GNN achieves the highest performance on the CICIoT2022 dataset, this non-pre-trained model falls significantly behind \name{} on the other datasets, highlighting its unstable classification performance.
% \subsubsection{CICIoT2022}
% Concerning the CICIoT2022 dataset, \name{} outperforms all baselines except TFE-GNN, which achieves a slight advantage of 0.15\% across all evaluation metrics. TFE-GNN represents traffic flow through a byte-level correlation graph and utilize graph neural networks to capture traffic patterns. However, TFE-GNN lags significantly behind \name{} on other datasets, particularly trailing by over 23\% on the ISCXTor2016 dataset, indicating its unstable classification performance.

% \subsubsection{ISCXTor2016, ISCXVPN2016 \& USTC-TFC2016}
% As depicted in \cref{tb:ablation}, \name{} surpasses all existing methods across the three encrypted datasets. Particularly noteworthy is the significantly unstable performance exhibited by methods other than YaTC and YaTC(OF) across different datasets. Given the similarities between \name{} and YaTC, we contend that a robust traffic representation scheme, which incorporates both header and payload data, alongside a well-designed pre-training task, plays a crucial role in enhancing encrypted traffic analysis capabilities.

\subsection{Inference Efficiency Evaluation}

% \begin{figure*}[htpb]
% \centering
% \includegraphics[scale=0.6]{figures/inference_efficiency.pdf}
% \caption{The Inference Speed and GPU Memory Comparison}
% \label{fig:efficiency}
% \end{figure*}

% \begin{figure}[htpb]
% \centering
% \includegraphics[scale=0.6]{figures/efficiency_scatter.pdf}
% \caption{The Inference Efficiency Comparison on Fine-tuning Batch Size}
% \label{fig:efficiency-bar}
% \end{figure}

\begin{figure*}[htpb]
    \centering
    \begin{minipage}{0.65\textwidth}
        \centering
        \includegraphics[scale=0.6]{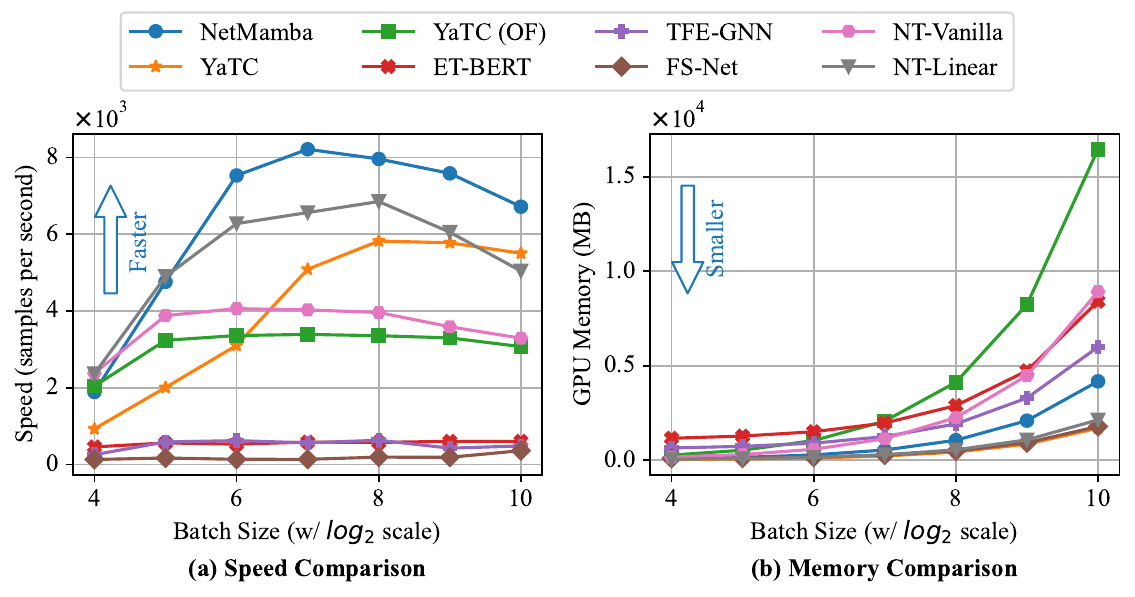}
        \caption{The Inference Speed and GPU Memory Comparison}
        \label{fig:efficiency}
    \end{minipage}
    \begin{minipage}{0.3\textwidth}
        \centering
        \includegraphics[scale=0.6]{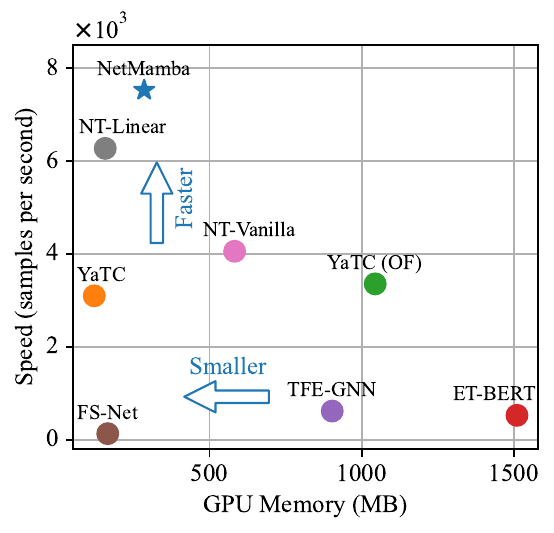}
        \caption{The Inference Efficiency Comparison on Fine-tuning Batch Size}
        \label{fig:efficiency-bar}
    \end{minipage}
\end{figure*}

To evaluate the inference efficiency of \name{}, we conducted experiments comparing its speed and GPU memory consumption with existing deep learning methods. Speed is measured as the number of traffic data samples processed by the model per second: 
packets for ET-BERT and flows for the others.
As shown in \cref{fig:efficiency}(a), \name{} achieves the highest inference speed across various input batch sizes, with improvements ranging from 1.22 to 60.11 times. 
This advantage is particularly notable due to the substantial model parameters and inefficient model architecture design present in models such as ET-BERT, TFE-GNN, and FS-Net.
Even when compared with models possessing similar parameter counts, \name{} continues to outperform NT-Vanilla, YaTC and its faster variant. This superiority is primarily attributed to Mamba's lower computational complexity compared to Transformer models. Given a token sequence $\mathbf{X} \in \mathbb{R}^{1\times \mathtt{L} \times \mathtt{D}}$ and the default setting $\mathtt{E} = 2\mathtt{D},\ \mathtt{N} = 16$, the computational complexities of vanilla or linear attention in Transformers and SSM in Mamba are as follows:
\begin{align}
    \Omega(\text{Vanilla-Attention}) = 4\mathtt{L}\mathtt{D}^2 + 2\mathtt{L}^{2}\mathtt{D} \\
    \Omega(\text{Linear-Attention}) = 3\mathtt{L}\mathtt{D}^2 + 2\mathtt{L}\mathtt{D} \label{eq:linear-attention} \\
    \Omega(\text{SSM}) = 3\mathtt{L}\mathtt{E}\mathtt{N} + \mathtt{L}\mathtt{E}\mathtt{N} = 96\mathtt{L}\mathtt{D} + 32\mathtt{L}\mathtt{D} \label{eq:ssm}
\end{align}

Self-attention exhibits quadratic complexity to the sequence length $\mathtt{L}$, whereas SSM operates linearly. This computational efficiency makes \name{} more scalable than Transformer-based models like YaTC and ET-BERT. Although both NT-Linear and \name{} achieve linear complexity and have similar parameter counts, based on \cref{eq:linear-attention} and (\ref{eq:ssm}), the SSM requires less computational cost when $\mathtt{D} > 42$. Given our use of $\mathtt{D} =256$, NT-Linear's slower inference speed compared to \name{} is reasonable.

In \cref{fig:efficiency}(b), \name{} demonstrates lower GPU memory consumption than most models, except FS-Net, YaTC and NT-Linear, when using large batch sizes. 
% FS-Net relies heavily on RNNs, which necessitate linear memory with respect to sequence length, allowing for reduced memory costs. However, it should be noted that FS-Net exhibits significantly slower inference speeds and poorer classification performance. 
FS-Net's reliance on RNNs, which require linear memory relative to sequence length, reduces memory costs but results in slower inference and poorer classification performance.
YaTC reduces memory usage by shortening input sequence length through a model forward trick. Without such an optimization, YaTC(OF) consumes up to four times more GPU memory than \name{}.
As shown in the subsequent ablation study, NT-Linear exhibits unstable classification performance due to over-compression of standard attention mechanisms.
Compared to other baselines, \name{} achieves improved memory efficiency primarily by customizing GPU operators that minimize the storage of extensive intermediate states and conduct recomputation during the backward pass.

When the input batch size is set to 64 (the value used in fine-tuning), as depicted in Figure \ref{fig:efficiency-bar}, \name{} exhibits an improvement in speed, being 2.24 times faster than the best baseline, YaTC(OF). Apart from FS-Net, memory-optimized YaTC, and NT-Linear, \name{} surpasses other methods in terms of GPU memory utilization. In summary, \name{} achieves the highest inference speeds among all deep learning methods while maintaining comparably low memory usage.

% In summary, \name{} achieves the fastest inference speeds among all deep learning methods, trailing only behind RNN-based FS-Net and memory-optimized YaTC in terms of GPU memory usage.

\begin{table*}[htpb]
    \caption{Ablation Study of \name{} on All Datasets}
    \begin{threeparttable}
    \begin{tabular}{c|cc|cc|cc|cc|cc|cc}
    \toprule
    \multirow{2}{*}{Method} & \multicolumn{2}{c|}{CrossPlatform(Android)} & \multicolumn{2}{c|}{CrossPlatform(iOS)} & \multicolumn{2}{c|}{CICIoT2022} & \multicolumn{2}{c|}{ISCXTor2016} & \multicolumn{2}{c|}{ISCXVPN2016} & \multicolumn{2}{c}{USTC-TFC2016}\\
    \cline{2-13} \noalign{\vskip 0.5mm}
    & AC & F1 & AC & F1 & AC & F1 & AC & F1 & AC & F1 & AC & F1 \\
    \midrule
    \textbf{\name{} (default)} & 
    \cellcolor{lightgray}0.9094 & \cellcolor{lightgray}0.9096 & \underline{0.9301} & \underline{0.9305} & 0.9928 & 0.9929 & \underline{0.9986} & \underline{0.9986} & \cellcolor{lightgray}0.9805 & \cellcolor{lightgray}0.9806 & \cellcolor{lightgray}0.9960 & \cellcolor{lightgray}0.9957 \\
    \midrule
    w/ Bidirectional Mamba\tnote{1} & 0.9012 & 0.9015 & 0.9213 & 0.9191 & \cellcolor{lightgray}0.9974 & \cellcolor{lightgray}0.9974 & 0.9966 & 0.9966 & 0.9704 & 0.9706 & \underline{0.9951} & \underline{0.9951} \\
    w/ Cascading Mamba\tnote{1} & 0.8194 & 0.8233 & 0.9015 & 0.8998 & 0.9687 & 0.9687 & 0.9952 & 0.9952 & 0.9320 & 0.9318 & 0.9852 & 0.9845 \\
    w/ Vanilla Transformer\tnote{2} &
    0.8836 & 0.8837 & 0.9058 & 0.9045 & \underline{0.9938} & \underline{0.9939} & 0.9973 & 0.9973 & 0.9632 & 0.9633 & 0.9954 & 0.9952 \\
    w/ Linear Transformer\tnote{2}
    & 0.6413 & 0.6857 & 0.4226 & 0.4717 & 0.8447 & 0.8558 & 0.8471 & 0.8708 & 0.7023 & 0.7361 & 0.7502 & 0.7742 \\
    w/o Position Embedding & \underline{0.9091} & \underline{0.9092} & 0.9103 & 0.9081 & 0.9872 & 0.9872 & 0.9979 & 0.9979 & 0.9588 & 0.9586 & 0.9920 & 0.9920 \\
    w/o Pre-training & 0.8868 & 0.8874 & 0.9151 & 0.9142 & 0.9882 & 0.9882 & 0.9966 & 0.9966 & 0.9335 & 0.9346 & 0.9904 & 0.9904 \\
    \midrule
    w/o Header & 0.5814 & 0.6626 & 0.7750 & 0.8124 & 0.5597 & 0.5318 & 0.8340 & 0.8434 & 0.5058 & 0.4394 & 0.5085 & 0.5672 \\
    w/o Payload & 0.9010 & 0.9004 & \cellcolor{lightgray}0.9365 & \cellcolor{lightgray}0.9362 & 0.9856 & 0.9856 & \cellcolor{lightgray}1.0000 & \cellcolor{lightgray}1.0000 & \underline{0.9747} & \underline{0.9747} & 0.9944 & 0.9945 \\
    % w/ IP Masking & 
    % 0.9049 & 0.9053 & \cellcolor{lightgray}0.9515 & \cellcolor{lightgray}0.9499 & 0.9918 & 0.9918 & 0.9973 & 0.9973 & 0.9559 & 0.9561 & \underline{0.9954} & \underline{0.9954} \\
    w/ Patch Splitting\tnote{3} & 0.8857 & 0.8863 & 0.9125 & 0.9103 & 0.9892 & 0.9892 & 0.9973 & 0.9973 & 0.9617 & 0.9619 & 0.9889 & 0.9890 \\
    \bottomrule
    \end{tabular}
    \begin{tablenotes}
        \item[1] Substituted unidirectional Mamba blocks with either bidirectional blocks \cite{zhu2024vision} or cascading ones\cite{chen2024mim}.
        \item[2] Replaced Mamba blocks with either vanilla \cite{vaswani2017attention} or linear\cite{katharopoulos2020transformers} Transformer blocks, termed NT-Vanilla and NT-Linear, respectively.
        \item[3] Changed the 1-dimensional stride cutting to 2-dimensional patch splitting.
    \end{tablenotes}
    \end{threeparttable}
    \label{tb:ablation}
\end{table*}

\begin{figure*}[h]
\centering
\includegraphics[scale=0.6]{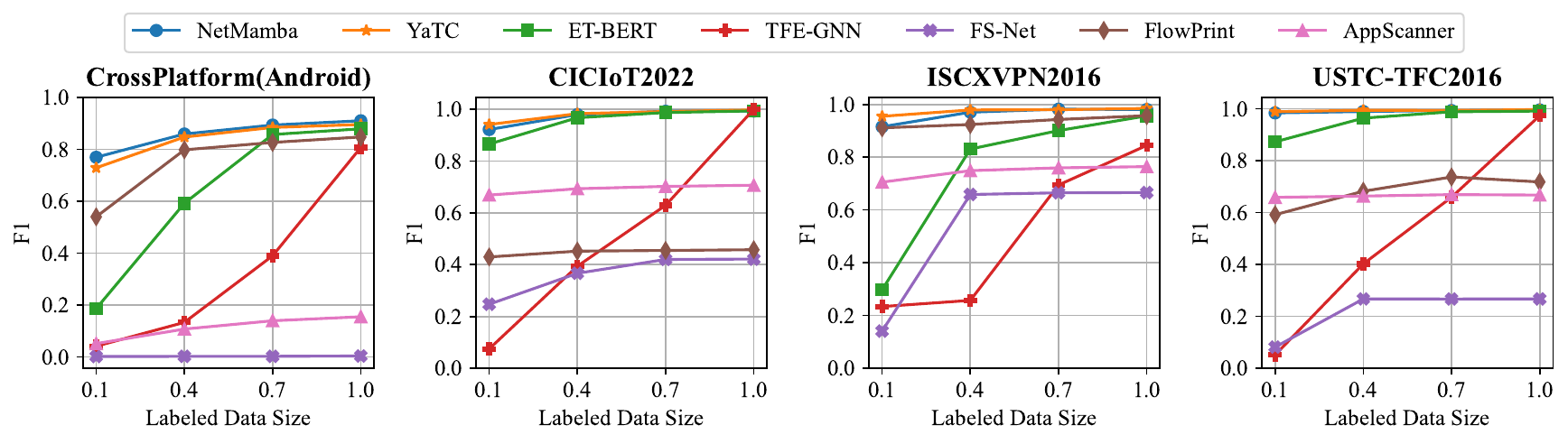}
\caption{The Performance Comparison on Few-Shot Settings}
\label{fig:few-shot}
\end{figure*}

\subsection{Ablation Study \label{sec:ab-study}}
To further validate both the model design and the traffic representation scheme of \name{}, we conducted ablation studies to assess the contribution of each component across six public datasets. The results are presented in \cref{tb:ablation}.

\subsubsection{Model-level Ablation} 
Initially, we replaced all unidirectional Mamba blocks in \name{} with bidirectional ones used by \cite{zhu2024vision}. The experimental results revealed a slight performance decline across all datasets except for CICIoT2022. This suggests that unidirectional Mamba is well-suited for processing network traffic data, given that packets are transmitted sequentially and earlier packets possess limited information about subsequent ones. Moreover, incorporating bidirectional or even omnidirectional Mamba blocks introduces additional computational and memory overheads due to extra scan passes, ultimately reducing efficiency. Thus, unidirectional Mamba stands out as the preferable choice.

Following \cite{chen2024mim}, we substituted the original unidirectional Mamba block with a cascading structure where each inner block processes data of different granularity. We observed a notable drop in classification performance across all datasets, indicating that this complex structure is inferior to processing sequential traffic data.

% To clarify the source of \name{}'s classification performance gains over existing Transformer-based models,
To further investigate the expressiveness of the Mamba architecture, 
we evaluated two Transformer-based ablation variants, NT-Vanilla and NT-Linear. The results show that NT-Vanilla performs slightly worse than \name{} across all datasets except for CICIoT2022. This indicates that the linear-time Mamba model is well suited for capturing sequential traffic data using our proposed representation scheme.
% With this representation scheme, a linear-time Mamba-based classifier achieves classification performance comparable to that of a quadratic-time Transformer-based model. 
Moreover, NT-Linear, due to information loss from the over-compression of its attention mechanisms, exhibits unstable performance and falls significantly behind \name{} on three datasets.

While positional information is inherently preserved in sequence models such as Mamba, eliminating explicit positional embedding still results in a reduction in accuracy ranging from 0.03\% to 2.17\% across all datasets. 
This suggests that reinforced positional information aids the model in capturing correlations within sequential traffic data.
% Therefore, incorporating positional information enhances the model's capability for better traffic comprehension.

The pre-training process is designed to capture general traffic understanding from extensive unlabeled data. When compared to the non-pre-trained counterpart, pre-trained \name{} demonstrates accuracy improvements ranging from 0.20\% to 4.70\%, affirming the effectiveness of our MAE-based pre-training task.

\subsubsection{Data-level Ablation}
When all header bytes in a packet are omitted, classification performance declines significantly, with accuracy dropping by 15.51\% to 48.75\%. This highlights the critical role of key fields within packet headers—such as port number, protocol, and packet length—which have been proven effective in traffic classification \cite{liu2019fs,madhukar2006longitudinal,fu2021realtime}.
% This underscores the importance of key fields within headers at the network, transport, and upper layers. 

Regarding packet payloads, the ablation results show accuracy drops ranging from 0.16\% to 0.84\% across four datasets.
% , with a slight increase of 0.07\% on the ISCXTor2016 dataset. 
This underscores the contribution of potential plaintext and specific encrypted payloads to improved traffic understanding.

% Without IP masking, the model may learn biased shortcuts based on IP addresses present in the training set, resulting in a maximum decrease in accuracy of 6.27\% in the test set.

Likewise, the vertical bias information introduced by the 2-dimensional patch splitting results in a maximum accuracy decline of 1.88\%, highlighting the importance of 1-dimensional stride cutting.

\subsection{Few-Shot Evaluation}
To validate the robustness and generalization abilities of \name{}, we conduct few-shot evaluations on four datasets, with labeled data size set to 10\%, 40\%, 70\%, and 100\% of the full training set (comprising 80\% of the total data). Specifically, we adopt a leave-one-out approach on the pre-training datasets to assess the transfer learning capability of the pre-trained models. In detail, the dataset used for fine-tuning is excluded from the pre-training datasets.
As shown in \cref{fig:few-shot}, the three pre-trained models, \name{}, YaTC, and ET-BERT generally outperform other supervised methods under few-shot and leave-one-out settings. While conventional machine learning methods like FlowPrint and AppScanner show some robustness to limited labeled data, their classification performance varies significantly across different datasets. Although the supervised TFE-GNN model performs comparably to the pre-trained models with the full training dataset, its performance drops considerably with smaller training data sizes. Thus, pre-trained models demonstrate superior robustness and generalization capabilities due to their ability to extract high-quality traffic representations from large amounts of unlabeled data, thereby reducing the dependence on labeled data.

Among the pre-trained methods, ET-BERT demonstrates lower reliability on two datasets, while YaTC performs comparably to \name{} on all datasets. Consequently, our model exhibits exceptional robustness, on par with Transformer-based models, and proves highly effective in addressing classification tasks with limited encrypted traffic data.

\section{Conclusion and Future Work}
In this paper, we introduce \name{}, a novel pre-trained state space model designed for efficient network traffic classification. 
To enhance model efficiency while maintaining performance, we utilize the unidirectional Mamba architecture for traffic sequence modeling and develop a comprehensive representation scheme for traffic data. 
Evaluation experiments on six public datasets demonstrate the superior effectiveness, efficiency, and robustness of \name{}.
Beyond classical traffic classification tasks, the comprehensive representation scheme and refined model design enable \name{} to address broader tasks within the network domain, such as quality of service prediction and network performance prediction. However, the current implementation of \name{} depends on specialized GPU hardware, which limits its deployment on real-world network devices. In the future, we plan to explore solutions to implement \name{} on resource-constrained devices.

% Initially, we pre-train \name{} on large unlabeled datasets using an MAE-based approach to capture generic network domain knowledge. We then fine-tune \name{} on limited labeled traffic data for various tasks such as encrypted application, attack, and malware traffic classification, ensuring adaptability across different scenarios. 

% We evaluate \name{} against state-of-the-art methods on six public datasets, demonstrating its superiority with up to 4.83\% higher accuracy and 4.64\% higher f1 score. Moreover, \name{} exhibits efficient performance, being 2.24 times faster than the best baseline while maintaining low memory usage. We conduct a comprehensive ablation study to validate \name{}'s model design and representation scheme, and verify its robustness and generalization through few-shot comparisons.

% \section{Acknowledgement}
% We thank Prof. Dario Rossi for his valuable suggestions and the anonymous reviewers for their insightful comments.
% This work is supported by the National Key R\&D Program of China (No. 2023YFB3106304).

% \clearpage
% \newpage

\bibliographystyle{IEEEtran}
\bibliography{reference}

\end{document}